\documentclass[10pt,letterpaper,compsoc,conference]{iiswc22}
\usepackage{cite}
\usepackage{amsmath,amssymb,amsfonts}
\usepackage{algorithmic}
\usepackage{graphicx}
\usepackage[dvipsnames]{xcolor}
\usepackage[final]{microtype}
\usepackage[italic]{mathastext}
\usepackage{libertine}
\usepackage[T1]{fontenc}
\usepackage{textcomp}
\usepackage[varqu,varl]{zi4}
\usepackage[all]{nowidow}
\usepackage[auth-lg,affil-it]{authblk}
\usepackage[keeplastbox]{flushend}
\usepackage{graphicx}
\usepackage{subfigure}
\usepackage{booktabs} 
\usepackage{mathtools}
\usepackage{comment}
\usepackage{algorithm}
\usepackage{multirow}
\usepackage{authblk}
\usepackage{amsmath,amssymb,amsfonts}
\usepackage{algorithmic}
\usepackage{textcomp}
\usepackage{xcolor}
\usepackage{xspace}

\usepackage{soul,color}  
\usepackage{authblk}
\usepackage{ragged2e}
\usepackage{makecell}
\usepackage[figurename=Fig.]{caption}
\usepackage[final]{microtype}
\usepackage{mathptmx}
\usepackage{fancyhdr}
\usepackage[normalem]{ulem}
\usepackage[hyphens]{url}
\usepackage[keeplastbox]{flushend}

\usepackage[bookmarks=true,breaklinks=true,letterpaper=true,colorlinks,citecolor=blue,linkcolor=blue,urlcolor=blue]{hyperref}

\usepackage{tcolorbox}

\newcommand{\PCignore}[1]{}

\def\Snospace~{\S{}}

\newcommand{\squishlist}{
 \begin{list}{$\bullet$}
  { \setlength{\itemsep}{0pt}
     \setlength{\parsep}{3pt}
     \setlength{\topsep}{3pt}
     \setlength{\partopsep}{0pt}
     \setlength{\leftmargin}{1.5em}
     \setlength{\labelwidth}{1em}
     \setlength{\labelsep}{0.5em} } }

\newcommand{\squishlisttwo}{
 \begin{list}{$\bullet$}
  { \setlength{\itemsep}{0pt}
     \setlength{\parsep}{0pt}
    \setlength{\topsep}{0pt}
    \setlength{\partopsep}{0pt}
    \setlength{\leftmargin}{2em}
    \setlength{\labelwidth}{1.5em}
    \setlength{\labelsep}{0.5em} } }

\newcommand{\squishend}{
  \end{list}  }

\newcommand{\TK}[1]{\textcolor{red}{TK: #1}}

\begin{document}

\renewcommand\Authsep{\qquad}
\renewcommand\Authand{\qquad}
\renewcommand\Authands{\qquad}

\title{Demystifying Map Space Exploration for NPUs } 

\author[1]{Sheng-Chun Kao}
\author[2]{Angshuman Parashar}
\author[2]{Po-An Tsai}
\author[1]{Tushar Krishna}

\affil[1]{Georgia Institute of Technology}  \affil[2]{Nvidia}
\affil[1]{\textit {skao6@gatech.edu},  \textit{tushar@ece.gatech.edu}}
\affil[2]{\textit { \{aparashar, poant\} @nvidia.com}}
\renewcommand\Authands{ and }

\maketitle

\begin{abstract}
Map Space Exploration is the problem of finding optimized mappings of a Deep Neural Network (DNN) model on an
accelerator. It is known to be extremely computationally expensive, and there has been active research looking at both heuristics and learning-based methods to make the problem computationally tractable. 
However, while there are dozens of mappers out there (all empirically claiming to find better mappings than others), the research community lacks systematic insights on how different search techniques navigate the map-space and how different mapping axes contribute to the accelerator's performance and efficiency. Such insights are crucial to developing mapping frameworks for emerging DNNs that are increasingly irregular (due to neural architecture search) and sparse, making the corresponding map spaces much more complex. In this work, rather than proposing yet another mapper, we do a first-of-its-kind apples-to-apples comparison of search techniques leveraged by different mappers. Next, we extract the learnings from our study and propose two new techniques that can
augment existing mappers --- {\em warm-start} and {\em sparsity-aware} --- that demonstrate speedups, scalability, and robustness across diverse DNN models\footnote{Code avaliable at https://github.com/maestro-project/gamma-timeloop.}.

\end{abstract}

\section{Introduction}

Deep Neural Network (DNNs) have become an indispensable tool in the solution toolbox for a variety of complex problems such as object detection, machine translation, language understanding, autonomous driving, and so on. There is growing demand for specialized DNN accelerators (also called Neural Processing Units or NPUs)\footnote{In this paper, we use the terms DNN Accelerator and NPU interchangeably.} pursuing high performance with high energy, power, and area efficiency.

\begin{figure*}
\begin{center}
\includegraphics[width=1\linewidth]{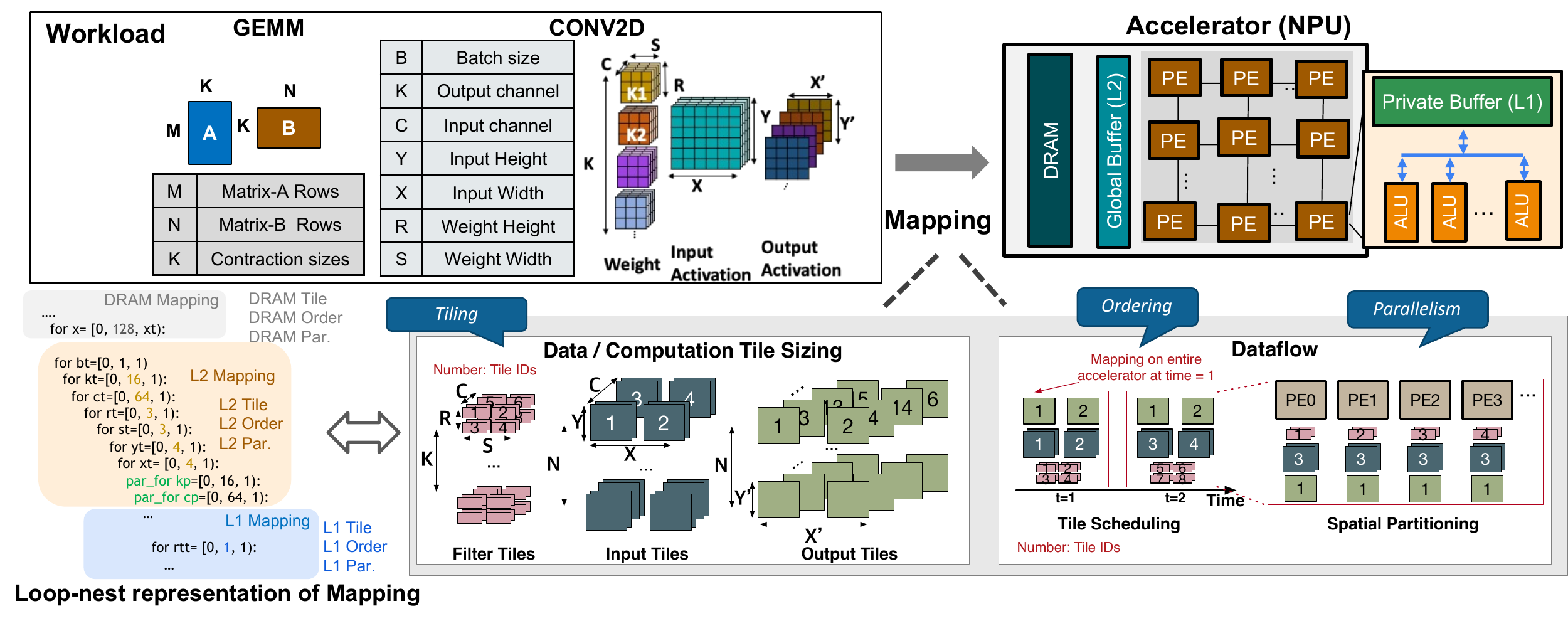}
\end{center}
\vspace{-0.40cm}
    \caption{The overview of DNN Workload, Accelerator, and a (NVDLA-like~\cite{nvdla}) Mapping.}

\label{fig:arch_map}
\end{figure*}

The performance and energy-efficiency of a NPU depends on how a DNN is \textit{mapped} over the accelerator's hardware (compute and memory) resources~\cite{timeloop, kwon2020maestro}. Specifically, 
a mapping (\textit{aka} schedule) includes the computation order, parallelization strategy and tile sizes~\cite{timeloop,kwon2020maestro}, as shown in \autoref{fig:arch_map}. In order to achieve high efficiency across a wide range of DNNs that include diverse layer shapes and sizes, state-of-the-art DNN accelerators are often designed with \textit{flexibility} to support different mapping strategies~\cite{eyeriss_v2,maeri,qin2020sigma}. 
This flexibility imposes a unique challenge for deployment: finding a high-quality mapping between a DNN and the flexible accelerator from the space of all legal mappings (i.e., the \emph{map space}) during compile time. 
This is crucial to 
unlock the full potential of the DNN accelerator.

As a result, prior work has clearly defined {\em map space exploration} (MSE)
\cite{timeloop,gamma,mindmapping,cosa}, as a critical problem for NPU design and/or deployment, cleanly separating it from the hardware architecture design space exploration (DSE) problem. 
DSE includes identifying the right compute and memory configurations for the NPU  within constraints such as total FLOPS, area, and power. MSE, meanwhile, takes the hardware configuration and DNN workload as input and finds optimized mappings, optimizing some objective (e.g., latency or energy-efficiency).
To perform MSE, various search algorithms (i.e., \emph{mappers}) have been proposed within the past few years~\cite{timeloop, simba, dmazerunner, yang2020interstellar,tillet2019triton, lu2017flexflow, gao2019tangram,tetris, deeptools, suda2016throughput, shen2017maximizing, cong_fpga, scaledeep, hypar,systolic_mapping, song2018towards,stoutchinin2019optimally,autotvm,hasco,reagen2017case,ahn2019reinforcement,flextensor,cosa,flexflow,ragan2013halide,vasilache2018tensor,baghdadi2019tiramisu,grosser2011polly}.


Despite the success achieved by these prior efforts, MSE remains a computationally challenging problem. 
This is because the search space for legal mappings for even a single layer of a modern DNN (e.g., ResNet-50) on a typical edge class accelerator~\cite{eyeriss_v2} is $\sim$ $O(10^{24})$~\cite{gamma,mindmapping} 
which would require more time than the age of the earth to search exhaustively (assuming 1ms to evaluate each mapping sample). This gets exacerbated as newer and ever larger DNN models are being created with increasing frequency, especially thanks to the success of neural architecture search techniques~\cite{mnasnet,cai2018proxylessnas,pham2018efficient,liu2018progressive,onceforall}. Furthermore, the advent of \textit{compressed-sparse} DNNs~\cite{wang2020spatten,guo2019reweighted,wang2019structured,sajjad2020poor, liu2018rethinking, li2016pruning, zhu2017prune}, whose mappings are not performance-portable across sparsity levels (a key finding in this paper), further increases MSE burden.


Researching more sophisticated scalable and sparsity-aware MSE techniques is at least partially hampered by the fact that 
even though prior approaches have \emph{empirically shown} that their techniques work, none of them
demonstrate \emph{why} they work and the insight behind their optimization techniques.
%

It is these very insights that we wish to extract in this paper, and in the process demystify MSE as a problem.
We cover both heuristics and learning-based optimization approaches, analyze their behavior, and learn from their best traits. We then use these learnings to scale MSE to more complex workloads.


Specifically, our contributions are two-fold.

(1) This is the first work, to the best of our knowledge, to \textit{quantitatively} compare three wide categories of mappers: random-based~\cite{timeloop} (i.e., heuristic pruning), feedback-based~\cite{gamma} (i.e., blackbox optimization and reinforcement learning), and gradient-based~\cite{mindmapping} (i.e., surrogate models), and analyze their trade-offs. We conduct a sensitivity analysis of different mapping axes to understand the contribution of each axis. We then perform case studies that reveal distinguishing characteristics of good and bad mappings. Our analysis reveals that: (i) random search is inefficient, (ii) gradient-based search converges fast but requires prior knowledge of the accelerator architecture, and (ii) feedback-based search is more adaptable and sample-efficient, but requires higher cost to acquire each sample.
Our analysis also shows that optimality of a dense DNN mapping does not port over to a sparse DNN.

(2) Based on our findings, we propose two novel heuristic techniques to advance the state-of-the-art in MSE:
(i) We propose a {\em warm-start} technique to initialize the MSE with
prior optimal
solutions from previous layers in
a replay buffer based on a \textit{similarity} metric, enabling the mapper to start at a better point and converge faster. In our evaluations, we find that warm-start can help the mapper converge to a similar performance point 3.3x-7.3x faster.
(ii) We also propose a {\em sparsity-aware} technique to search for a mapping that can perform well across a range of target activation sparsities. A fixed mapping found by our sparsity-aware approach can achieve 99.7\% of the performance of each of the mappings specifically tailored to the various density levels.

\section{Background: DNN Accelerators}
\label{sec:background}
\vspace{-1mm}

\subsection{DNN Workloads}
\label{sec:background_workload}
\vspace{-1mm}
In this work, we use individual DNN layers/operators as our target \textit{workload}.
The workloads vary across different DNN models because of different types of operations such as CONV2D, Depth-wise CONV, Point-wise CONV, Attention, Fully-Connected (FC), and so on, and different tensor shapes for the layers (i.e., batch, input, weight kernel sizes), as shown in  \autoref{fig:arch_map}. All these operations can be represented with a loop-nest of computations. For example, a CONV2D can be represented as 7 for-loops, and GEMM can be represented as 3 for-loops.

\subsection{Accelerator Hardware Configuration}
\label{sec:hw_configuration}
\vspace{-1mm}
A canonical NPU often houses a spatial array of Processing Elements (PEs), as shown in \autoref{fig:arch_map}. Each PE has one to several ALU units to compute partial sums, and private local (aka ``L1'') buffers to store weights, input activations
and partial sums. The accelerator also houses a global shared (aka ``L2'') buffer to prefetch activations and weights from DRAM for the next tile of computation that will be mapped over the PEs and L1 buffers. Networks-on-Chip are used to distribute operands from the global L2 buffer to the L1 buffers in the PEs, collect the partial or full outputs, and write them back to the L2 buffer. 

\begin{figure*}[t]
\begin{center}
\includegraphics[width=0.8\linewidth]{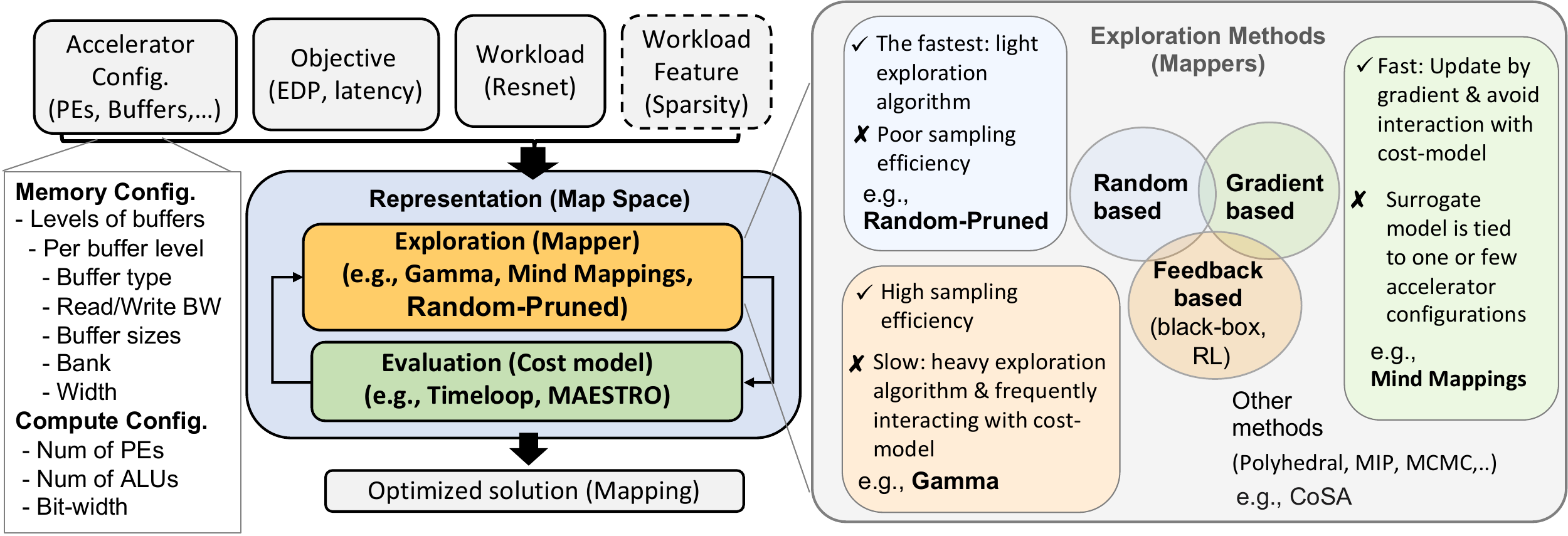}
\end{center}
\vspace{-0.40cm}
    \caption{A canonical Map Space Exploration framework.}
\vspace{-0.2cm}
\label{fig:mse}
\end{figure*}

\vspace{-1mm}
\subsection{Accelerator Map-Space}
\label{sec:map_space}
\vspace{-1mm}
Given a DNN workload, there exist several 
choices for \emph{mapping} it on the accelerator's PEs and buffer hierarchy over space and time. The mapping includes the following components~\cite{maestro, timeloop}, shown in \autoref{fig:arch_map}:


\textbf{(1) Tile sizes:} 
The ability to change bounds and aspect ratios of data tiles from one or more operand tensors per level of the buffer hierarchy~\cite{buffets}.

\textbf{(2) Loop order:} The ability to change the loop orders iterated per tiling level.

\textbf{(3) Loop parallelization:} The ability to change which tensor dimensions are parallelized per tiling level. This represents the \textit{spatial} partitioning of data (i.e., across PEs).

\autoref{fig:arch_map} shows an example of the mapping used by the NVDLA~\cite{nvdla} accelerator.
Choices for (2) and (3) together are often referred to as \emph{dataflow}~\cite{maestro} which has been informally classified by prior work into weight-stationary, output stationary and input-stationary~\cite{chen2016eyeriss}. The design-space of all possible mappings (i.e., dataflows + tile-sizes) that an accelerator can support is called its \emph{Map-Space}~\cite{timeloop}.

\textit{Flexible} DNN accelerators~\cite{maeri, eyeriss_v2} allow a mapping optimizer within a compiler to explore tile sizes, loop orders and parallelization independently for each layer. This mapping flexibility is crucial for accelerators to adapt to growing diversity in DNNs~\cite{maestro}.
The overall runtime and energy-efficiency of an accelerator depends on both the hardware configuration and the mapping, making it crucial to find an optimized mapping\footnote{In this paper, we focus on finding optimized mapping for individual DNN layers/operators, which has been the target of most Map-Space Exploration tools. We leave Inter-layer mappings via operator-fusion as future work.},
~\cite{timeloop, maestro, yang2020interstellar}, as we discuss next.

\section{Map Space Exploration (MSE)}
\vspace{-1mm}

A canonical MSE framework is shown in \autoref{fig:mse}.
MSE takes the NPU's HW configuration (\autoref{sec:hw_configuration}) and target DNN workloads (size, shape, and additional features such as sparsity level of weight and/or activations) as input and finds optimized mappings given 
an objective (e.g., latency, throughput, energy, energy-delay-product (EDP), and so on).
MSE may be run at compile time within a mapping optimizer~\cite{marvel} after the NPU is deployed, or at design-time in conjunction with DSE for
co-optimizing the mapping and HW configuration~\cite{hasco, digamma}.

The MSE process often includes three parts: \emph{Representation} of search space, \emph{Evaluation method}, and \emph{Exploration method}. The representation will define the scope of the searching problem and the size of the search space. 
An optimization loop that includes exploration and evaluation performs the actual search.
The optimization continues till the MSE converges, or reaches a given sampling budget or wall-clock run time budget.

\subsection{Representation of Map Space}
\vspace{-1mm}
While recent work has proposed various representations (MAESTRO~\cite{kwon2020maestro}, UNION~\cite{jeong2021union}, and Ruby~\cite{horeni2022ruby}) to increase mapping diversity in the map space, in this work we leverage the canonical Timeloop representation, which is loop-nests to represent each tiling level (e.g., NVDLA-like mapping in \autoref{fig:arch_map}). We ensure that all the candidate mappings generated by various mappers during MSE are legal. 

\subsection{Evaluation Method (Cost Model)}
\label{sec:cost_model}
\vspace{-1mm}
MSE relies on a DNN accelerator \emph{cost model} to estimate the performance of a certain mapping on a given accelerator for a given workload. 
These cost models are typically analytical, enabling rapid evaluation of different design-points in a matter of ms. 
Some widely used cost models include Timeloop~\cite{timeloop}, MAESTRO~\cite{maestro}, dMazeRunner~\cite{dmazerunner}, Interstellar~\cite{yang2020interstellar}, SCALE-sim~\cite{scalesim} and others~\cite{zigzag, flat}. These cost models can model different kinds of accelerators (systolic arrays~\cite{scalesim}, flexible spatial arrays~\cite{maestro, timeloop, dmazerunner}, sparse accelerators~\cite{wu2021sparseloop}, and so on) and capture each accelerator's map space in different formats.
%
In this work, we use Timeloop~\cite{timeloop} as our cost model\footnote{Timeloop includes \textit{both} a cost model and mappers. Throughout this paper, we refer to the former as Timeloop and the latter as Timeloop-mapper. Timeloop-mapper itself supports a variety of search heuristics, with the default being Random-Pruned which we use. We also run other mappers using Timeloop as the cost model.} which is validated against real chips~\cite{eyeriss_isscc, simba}.


\subsection{Exploration Method (Mapper)}
\label{sec:mapper_intro}
\vspace{-1mm}
The exploration algorithm in MSE (\autoref{fig:mse}) is called a mapper. Dozens of different DNN mappers have been proposed, which we categorize into \textit{random search based}~\cite{timeloop, simba, dmazerunner, yang2020interstellar,tillet2019triton}, \textit{feedback-based} (including reinforcement learning and black-box optimization) ~\cite{gamma, autotvm, hasco,confuciux, flexflow, flextensor}, \textit{gradient-based}~\cite{mindmapping}, and \textit{others} (including mathematical optimization, MCMC, polyhedral transformations, and heuristics)~\cite{cosa, flexflow,ragan2013halide,vasilache2018tensor,baghdadi2019tiramisu,grosser2011polly} (\autoref{fig:mse}). 
The random search-based either apply random sampling on the search space or apply pruned random search~\cite{timeloop, marvel}, which prunes off the redundant search space to increase the sampling efficiency. The feedback-based use a learning algorithm to interact with the cost model and keep improving its solution. The run time of both random search-based and feedback-based depend heavily on the run time of the cost model, potentially becoming the bottleneck of the MSE run time. Gradient-based methods uses a \textit{differentiable} surrogate model, which eliminates this bottleneck and can update the solution directly by the gradient of the loss. We do a deeper dive within these three types in \autoref{sec:mapper_cpr}.


\subsection{Why MSE Matters}
\vspace{-1mm}
MSE bridges the gap between two active trends: (1) efficient DNN model design~\cite{sandler2018mobilenetv2,tan2019efficientnet,sparsetransformer} (which has led to a huge diversity in layer shapes/sizes and emergence of sparsity in state-of-the-art DNN models) and (2) flexible hardware accelerators that support diverse mappings (dataflows + tile sizes) via configurable buffer hierarchies~\cite{buffets} and on-chip interconnect topologies~\cite{maeri,qin2020sigma} as an answer to the first trend. MSE is crucial for extracting performance and energy-efficiency from the accelerator as there can be multiple orders of of difference in performance and energy-efficiency between good and bad mappings, as prior works have demonstrated~\cite{timeloop, gamma, mindmapping}.

\textit{While several mappers are being actively developed~\cite{timeloop, simba, dmazerunner, yang2020interstellar,tillet2019triton, lu2017flexflow, gao2019tangram,tetris, deeptools, suda2016throughput, shen2017maximizing, cong_fpga, scaledeep, hypar,systolic_mapping, song2018towards,stoutchinin2019optimally,autotvm,hasco,reagen2017case,ahn2019reinforcement,flextensor,cosa,flexflow,ragan2013halide,vasilache2018tensor,baghdadi2019tiramisu,grosser2011polly}, there is no work, to the best of our knowledge, that has focused on understanding how different mappers navigate the map-space, how different mapping axes contribute to the performance, and trade-offs between search approaches, which is the focus of this work.}

\section{Quantitative MSE Analysis}
\label{sec:analysis}

In this section, we perform a quantitative analysis of the three classes of mappers described in \autoref{sec:mapper_intro} to identify
\emph{when} and \emph{why} one works better than the other. The goal of this analysis is to educate the DNN accelerator research community on Mapper design, rather than propose yet another mapper.

\subsection{Methodology}
\label{sec:methodology}
\vspace{-1mm}

\textbf{Workload.} 
We consider workloads from different models: Resnet~\cite{resnet}, VGG~\cite{vgg}, Mnasnet~\cite{mnasnet}, Mobilenet~\cite{sandler2018mobilenetv2}, and Bert-large~\cite{transformer}. Some frequently referenced workloads across different experiments are described in \autoref{table:workload_table}.

\textbf{Hardware Accelerator.}
We model the NPU using Timeloop~\cite{timeloop}. We assume three-levels of buffer hierarchies: DRAM, a 64KB shared global buffer, and 256B private local buffer for each of the 256 PE. Each PE houses 4 ALU units (Accel-B in \autoref{table:workload_table}). We also model the NPU the Mind Mappings paper~\cite{mindmapping} uses (Accel-A), whose configuration is similar but with different sizing as shown in \autoref{table:workload_table}.

For analyzing sparse mappings (\autoref{sec:understanding_sparse}), we use TimeloopV2, \textit{aka Sparseloop}~\cite{wu2021sparseloop,wu2022sparseloop}, as the cost model to explore the map space in a flexible sparse accelerator, and leverage Gamma as the mapper. 
Besides tiling, orderering and parallelism, Sparseloop also models hardware and software optimizations (e.g., power gating and compressed tensors) in sparse DNN accelerators.

\textbf{Objective.}
We use multi-objective -- Energy and Latency (Delay), throughout the optimization process. When optimization finishes, we select the solution with the highest Energy-Delay-Product (EDP) on the Pareto frontier. We use EDP as the performance criteria of found mapping. Note that any formulation of the objective can also be used such as power, area, performance-per-watt, performance-per-mm$^2$, and so on. 

\textbf{Experiment Platform.}
We run experiments using a desktop with a 12-core Intel I7-6800K CPU and a Nvidia GTX1080 to train the surrogate model in Mind Mappings.

\begin{table}[t]

\centering
\caption{The description of the relevant workloads and accelerator configurations used across evaluations.}

\includegraphics[width=1\linewidth]{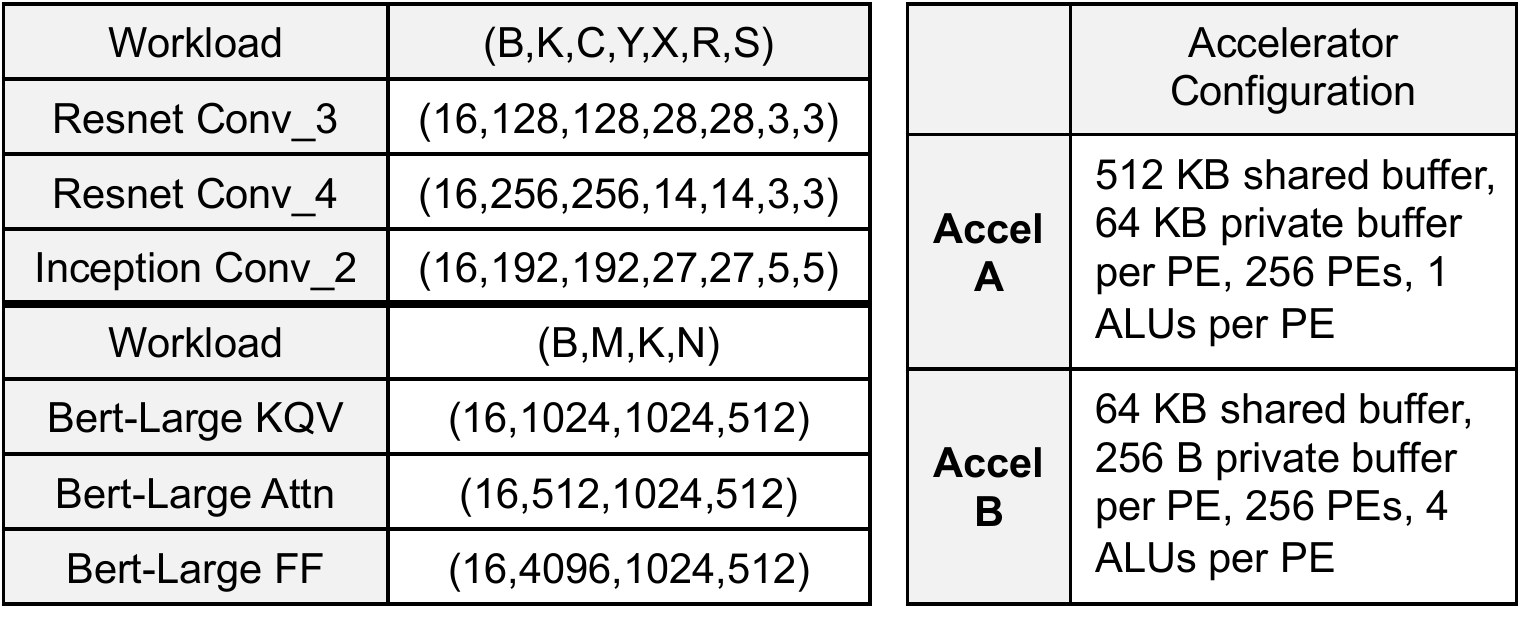}

\label{table:workload_table}
\end{table}

\begin{figure*}
\begin{center}
\includegraphics[width=0.9\linewidth]{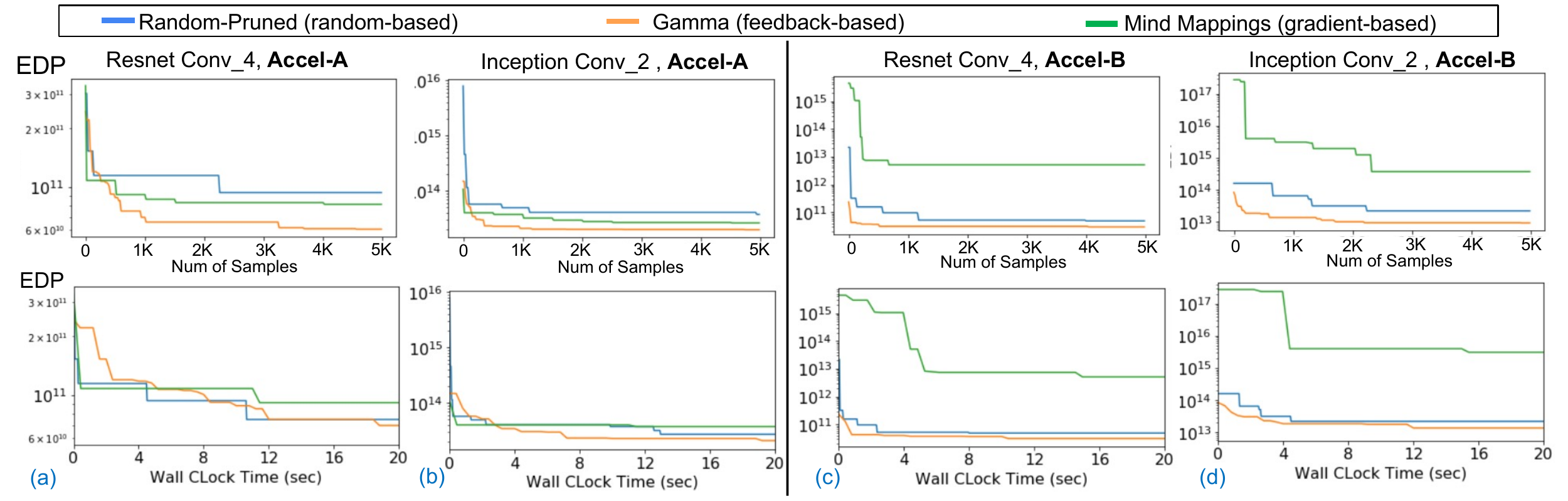}
\end{center}
\vspace{-0.50cm}
\caption{Comparisons of different types of mappers. Top figures show the converge curve across number of samples. Bottom figures show the converge curve across wall clock time.}
\vspace{-0.10cm}
\label{fig:mapper_cpr}
\end{figure*}

\begin{figure}[t]
\begin{center}
\includegraphics[width=1\linewidth]{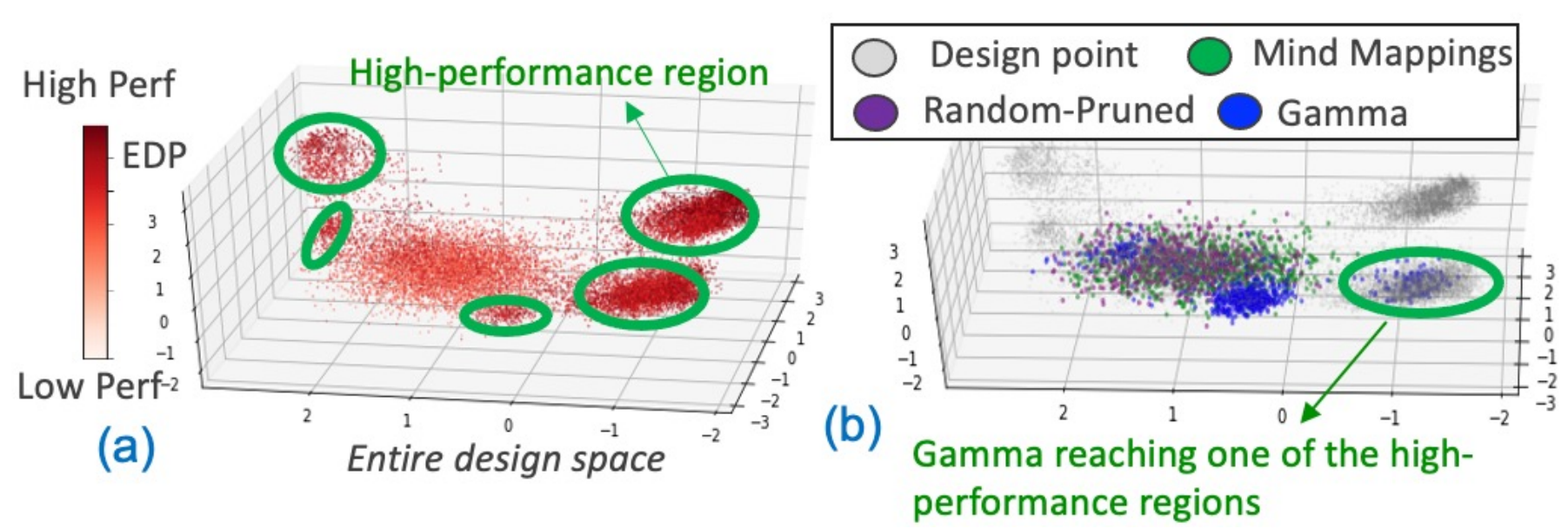}
\end{center}
\vspace{-0.4cm}
\caption{(a) shows the sampled points by exhaustively sampling the search space of (Resnet Conv\_4, Accel-A). The 3D visualization is projected by PCA dimension reduction. (b) shows the sampled points of different types of mappers in this search space.}
\label{fig:3d_plot}
\end{figure}

\subsection{Size of Map Space}
\label{sec:mapspace_size}
\vspace{-1mm}
The size of the map space heavily depends on representation. In this paper, we follow the efficient representation used by Timeloop 
to represent the three mapping axes. We use CONV2D (7 for-loop) as workload and 3-level of buffer hierarchy (DRAM, L2, L1) as architecture configuration as an example to guide the discussion of map space.

\textbf{Tile sizes.}
Buffers at each level of the scratchpad memory hierarchy will have a dedicated tile size for each of the dimensions, as shown by the different tile sizes within the 7 for-loops of the L2 mapping in \autoref{fig:arch_map}
The total possible combination depends on the tensor shape of each workload and increases exponentially with the number of buffer hierarchies.

\textbf{Loop Order.}
Each buffer level would have a dedicated permutation of loop order. E.g., in \autoref{fig:arch_map},
the loop order in L2 mapping from outer to inner loop is (B,K,C,R,S,Y,X). The total combinations become $(7!)^{3}$ (we have 3 buffer levels in our example).

\textbf{Parallelism.}
Parallelism happens across levels of compute units (2-level of compute units in \autoref{fig:arch_map},
i.e., across PEs and ALUs). At each level of the compute unit, we can choose to parallelize from 0 (no parallelism) to 7 (all parallelism) dimensions. The total combination becomes $2^{7 \times 2}$.

\textbf{Map-Space.}
The Cartesian product of these sub-spaces leads to the size of the entire map space, which is at the level of $O(10^{21})$ for the workloads discussed in \autoref{sec:methodology}.


\subsection{Understanding Mapper Sampling Efficiency}
\label{sec:mapper_cpr}
\vspace{-1mm}

Recall from \autoref{sec:mapper_intro} that we categorize state-of-the-art mappers into three major techniques (\autoref{fig:mse}). We select state-of-the-art mappers out of each category - Timeloop's Random-Pruned~\cite{timeloop} from random-based, Gamma~\cite{gamma} from feedback-based, and Mind Mappings~\cite{mindmapping} from gradient-based methods\footnote{Random-Pruned and Mind Mappings both natively work with the Timeloop cost model. Gamma was originally demonstrated with MAESTRO, and we extended it to use the Timeloop cost model. We leave the task of porting representative mappers from the \textit{others} category (\autoref{sec:mapper_intro} to a common cost model and analyzing them as future work.}. - and compare their characteristics with respect to search speed and sampling efficiency\footnote{The performance improvement over number of sampled points.}.

\squishlist
\item \textbf{Random-Pruned (random-based):} Random-Pruned~\cite{timeloop} uses random sampling on a pruned search space. The pruning strategies are based on heuristics, e.g., permutations do not matter for the innermost tiling level and for tile sizes that are one~\cite{timeloop}.

\item \textbf{Gamma (feedback-based):} Gamma~\cite{gamma}, a genetic algorithm (GA) based method, keeps a population of candidate solutions, uses specifically designed mutation operators to perturb populations to explore different mapping axes (tile, order, parallelism), and uses crossover to create next generations of populations. Gamma has been shown to beat other optimization techniques, including reinforcement learning~\cite{gamma,magma}.

\item \textbf{Mind Mappings (gradient-based):}
Mind Mappings~\cite{mindmapping} trains a neural-network-based surrogate model via offline sampling of millions of data points collected from the cost model. It uses the loss gradient to update its solution. During MSE, it utilizes gradient-descent on this surrogate model to find mappings, instead of searching.
\squishend

In the following evaluation case study, we show two sets of NPU configurations (\autoref{table:workload_table}) : \textit{Accel-A, on which the surrogate model is trained for MindMappings, and Accel-B, an unseen accelerator configuration for the surrogate model.}

\begin{figure*}
\begin{center}
\includegraphics[width=0.9\linewidth]{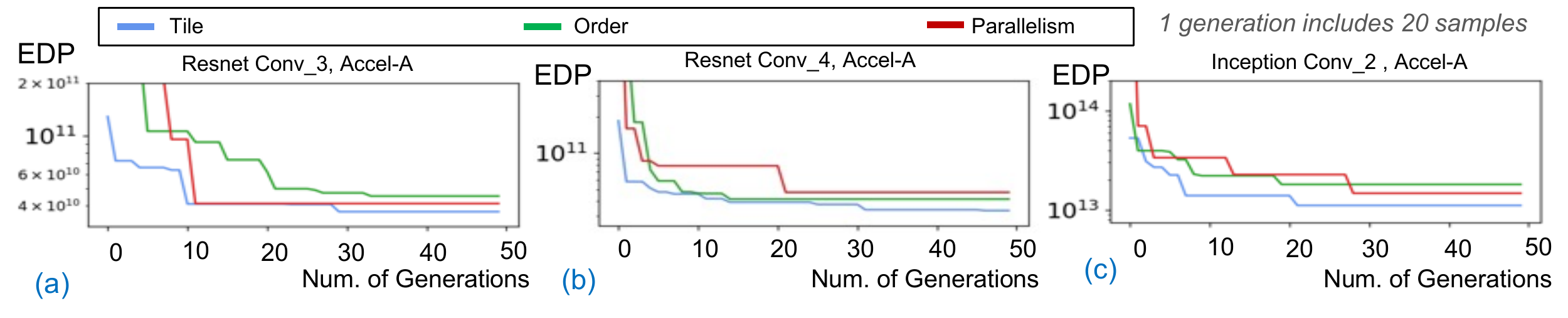}
\end{center}
\vspace{-0.50cm}
\caption{Mapping axes sensitivity analysis using the mutation operators in Gamma~\cite{gamma}. E.g., Tile (blue): means mutating tile only, i.e, only tile is explored, and other mapping axes are fixed, similarly for (mutate-)Order and (mutate-)Parallelism.}
\label{fig:operator_exp_single}
\end{figure*}

\begin{figure*}
\begin{center}
\includegraphics[width=0.9\linewidth]{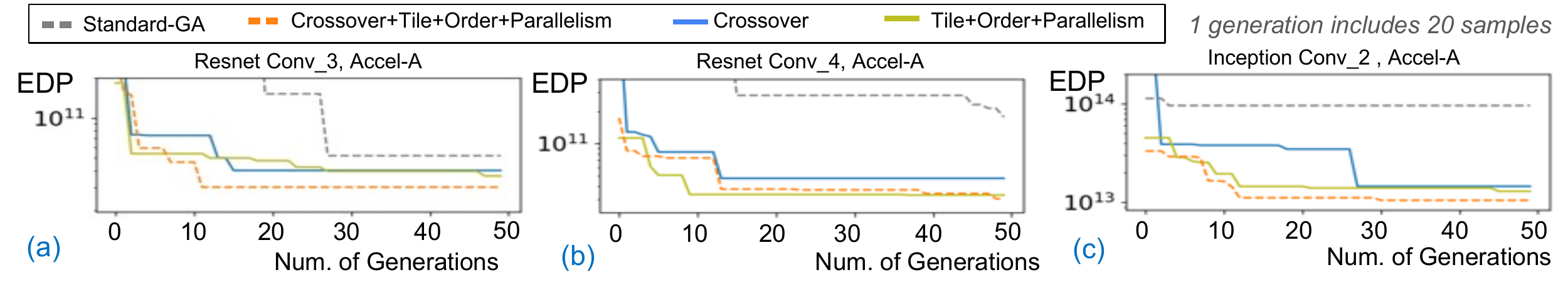}
\end{center}
\vspace{-0.50cm}
\caption{Crossover (blending two mappings) sensitivity analysis using operators in Gamma~\cite{gamma}. Standard-GA uses the standard mutation and crossover (without domain-specific operators along each mapping axes designed in Gamma~\cite{gamma}).}
\vspace{-0.10cm}
\label{fig:operator_exp_multiple}
\end{figure*}

\subsubsection{Trained Accelerator Configuration (Accel-A)}

\textbf{Iso-sampling points Comparisons.}
We set the sampling budget to 5,000 points and compare the sampling efficiency of algorithms in the top figures of \autoref{fig:mapper_cpr}(a)(b). The random-based method progresses the slowest over number of samples. Among the gradient-based and feedback-based, the gradient-based method progresses faster at the start owing to its direct gradient feedback. However, with more number of samples, the feedback-based method starts to perform better. It is because the gradient-based method is more prone to fall into local optimum (discussed later) while the feedback-based methods typically work well for global optimization problems. 

\textbf{Iso-time Comparisons.} 
We set a tight time budget, 20 seconds, and track the performance to wall clock time in the bottom figures of \autoref{fig:mapper_cpr}(a)(b). Despite their better sampling efficiency, the feedback-based and gradient-based methods do not show a clear edge over the random-based method within tight wall-clock run time budget. Random-based methods do not have costly built-in learning algorithms as the other two and hence can run more number of samples given the same time budget, which is essential when the run time budget is strictly tight. Specifically, the run time of the searching algorithm in Gamma and Mind Mappings is about 10x larger than Random-Pruned.

\subsubsection{Accelerator configuration not in the Training Dataset (Accel-B)}
We use the same set of workloads as in \autoref{fig:mapper_cpr}(a)(b), but change the accelerator configuration to Accel-B, which is not in the training dataset of the surrogate model of the gradient-based method. As shown in \autoref{fig:mapper_cpr}(c)(d), the gradient-based method cannot perform as well as it did for the trained accelerator configuration, Accel-A. It demonstrates that the trained surrogate model does not generalize across accelerator configurations. Note that we can also re-train the surrogate model for the new accelerator configuration, which will recover the performance. However, it will require another full-fledged DNN training. 
Besides, we also need to collect 1 - 5 million of new training data to achieve quality results~\cite{mindmapping}.

\textbf{Variance of Accelerator Configurations.}
The random-based and feedback-based method take workloads and accelerator configurations as inputs and therefore are agnostic to variance in accelerator configurations. In contrast, the gradient-based method train its surrogate model based on a collected training dataset. The training dataset includes collected workloads and collected accelerator configurations. While surrogate model can generalize the workload encoding across different DNNs models~\cite{mindmapping}, the generalization of accelerator configurations is more challenging since arbitrary buffer levels, buffer sizes, PE sizes, and other details (\autoref{fig:mse}) can be made. Thus the surrogate model is tied to one or few accelerator configurations.

\subsubsection{Visualization of the Sampling Points}
To better understand how different algorithms behave in the map space, we plot their sampling points in \autoref{fig:3d_plot} using the workload and accelerator configuration in \autoref{fig:mapper_cpr}(a). \autoref{fig:3d_plot}(a) shows the entire map space 
while dark red represent higher-performance points. There is a large low-performing region at the center while some small clusters of the high-performing points (green circle) scatter across the space.
\autoref{fig:3d_plot}(b) shows the points different algorithms actually sampled. Given the limited 5,000 sampling budget, The Random-Pruned method only samples around the lower-performing region because most of the design points sit here. Mind Mappings starts with the lower-performing region and gradient-updates to the higher-performing regions at the right. However, it sits at the local optimum. Gamma also starts with a lower-performing region but can explore a wider region faster because of its population-based method (which is common in many feedback-based algorithms~\cite{ga, tbpsa, pso_paper, cma}). Gamma reached one of the high-performance regions, as shown in \autoref{fig:3d_plot}(b).

\textbf{Takeaway of comparing different mappers:}
\squishlist

\item Learning-based methods, including gradient-based and feedback-based, can keep improving the quality of the sampling function over searching iterations, leading to better sampling efficiency.

\item When the time constraint is strictly tight so that the learning-based methods cannot yet gather adequate data to improve their sampling function (i.e., still at exploration phase instead of exploitation), the random-based method is the most cost-effective choice.

\item The surrogate model of the gradient-based method is trained on a collected training dataset, where the accelerator configuration is often fixed. The trained surrogate model cannot generalize across different accelerator configurations.
\squishend

\textit{We pick Gamma, the feedback-based method, as our main mapper for the rest of the discussion in this paper.}

\subsection{Understanding Mapper Search Operators}
\label{sec:ablation}

Recall that there are three mapping axes in the map space, tile, order, and parallelism. Gamma has dedicated genetic operators to explore along these axes, i.e., \textit{mutate-tile}, \textit{mutate-order}, and \textit{mutate-parallelism}. It also houses a \textit{crossover} operator to blend two high-performant mappings to create the next candidate mapping samples. Note that each genetic operator is specifically tuned to adapt to this map space as shown in the Gamma paper~\cite{gamma}, which is the key source of sampling efficiency over other black-box optimizers, including RL and standard GA. As \autoref{fig:operator_exp_multiple} shows, full-fledged Gamma (dotted orange line) performs an order of magnitude better than standard GA across the three evaluated workloads.

\subsubsection{Mapping Axis Sensitivity Analysis}
In \autoref{fig:operator_exp_single}, we explore each mapping axis individually (keeping the other two fixed) via the mutation operator in Gamma~\cite{gamma} such as mutate-tile for tile exploration, mutate-order for order exploration and so on.
We find mutate-tile to have the highest impact on EDP compared to the other components. 

\subsubsection{Crossover Sensitivity Analysis}
Gamma has crossover operator which blends two mapping points to create the next candidate mapping points. We execute a sensitivity analysis of crossover in \autoref{fig:operator_exp_multiple}. We find that disabling crossover (light green) can hugely impact the potential performance compared to full-fledged Gamma (dotted orange). However, crossover-only without other operators (dark blue) is also not adequate. 
Crossover working with all the dedicated mutation operators for the three maxing axes (dotted orange) can maximize the sampling efficiency of the mapper (Gamma) and ends up giving the most optimized performance.

\textbf{Takeaway of comparing operators in a mapper:}

\squishlist
\item If one were to incrementally implement different exploration functions along the mapping axes, starting with the tile exploration would be the most cost-effective option.

\item Blending two high-performance mappings (crossover) can effectively create another high-performance mapping.

\item The ability to explore different order and parallelism dimensions choices is not as critical as tile size exploration to optimize EDP performance. 

\item Note that even when fixing the order or parallelism throughout the optimization process, at the initialization stage, we still randomly initialized order and parallelism for the initial populations (a groups of initial sampling points). It implies that few explorations of order and parallelism are often adequate to give competitive mapping. It is owing to the fact that many combinations of order or parallelism will lead to similar latency or energy performance, as we discuss later in \autoref{sec:order_study}.

\item The performance difference of two mapping for the same problem can be as large as 3 orders of magnitude difference, consistent with prior works~\cite{gamma, maestro, timeloop, mindmapping}.

\squishend

\begin{figure}[t]
\begin{center}
\includegraphics[width=1\linewidth]{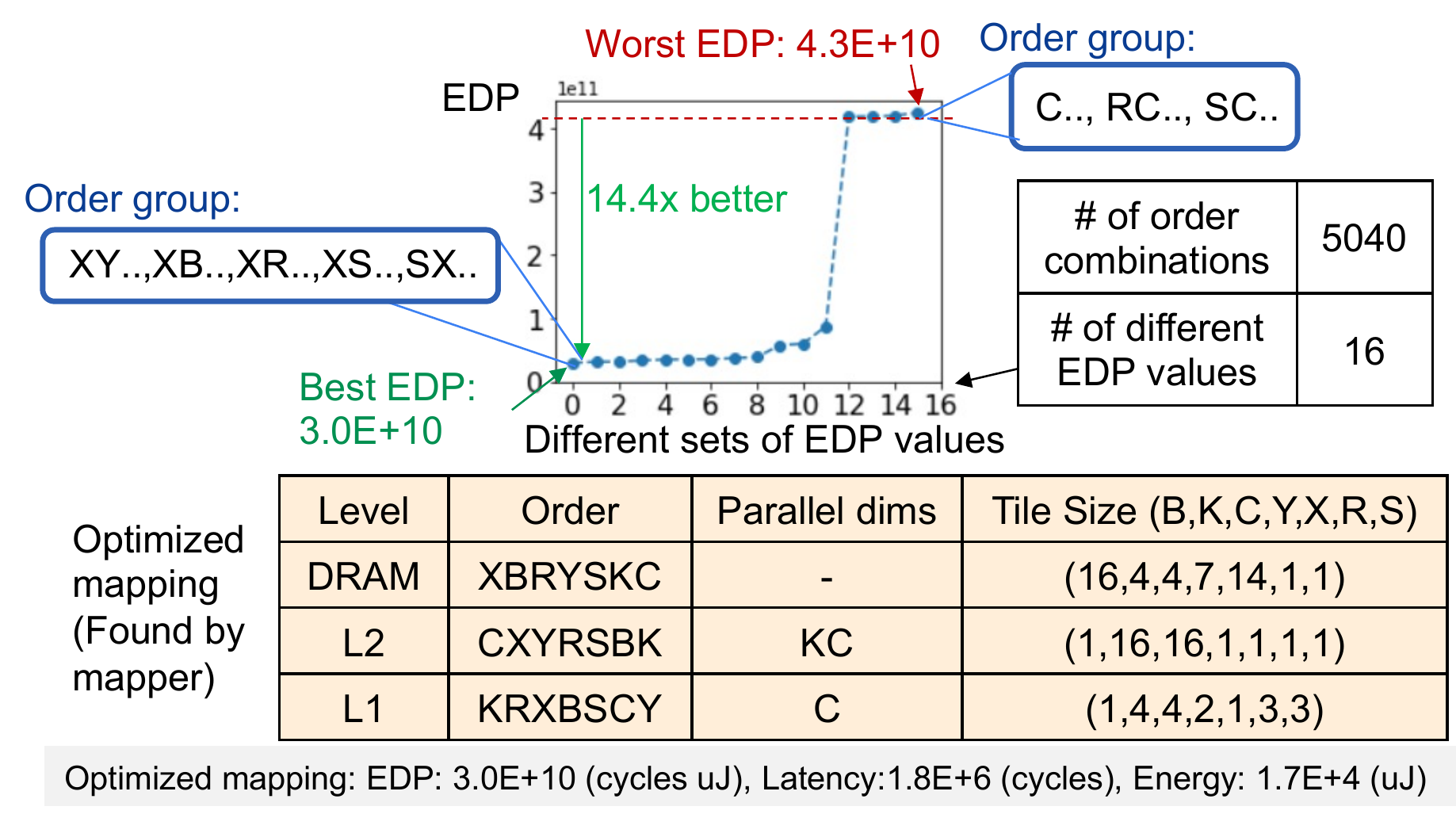}
\end{center}
\vspace{-0.40cm}
\caption{ The EDP difference of the same mapping with different loop order. We sweep through all 7! order combinations assuming all the buffer level utilize the same order. The 7! different mapping leads to 16 different EDP performance, with the best and the worst EDP differs by 14.4x times (under Resnet Conv\_4, Accel-B).}

\label{fig:order_exp}
\end{figure}

\subsubsection{Loop Order Sensitivity Analysis}
\label{sec:order_study}
We perform a sweep of loop order permutations to demonstrate our observation that \textit{many order permutations lead to similar performance} as observed above. We use the found mapping in the experiment setting in \autoref{fig:operator_exp_multiple}(a) and swap out the order permutation by enumerating through all the possibilities. The search space is as large as $(7!)^3$=1.28E+11. We add a constraint that each level of the buffer will use the same order to relax the complexity, which becomes 7!=5,040 choices. \autoref{fig:order_exp} shows that there are only 16 different EDP values out of 5,040 different mappings. We can observe some patterns in each of the same performance mapping groups, as shown in \autoref{fig:order_exp}. For example, ``XY..'' means the permutation starting with XY. The loop order at the DRAM buffer level of the original mapping found by Gamma (XB..) also falls in the high-performance order group.

\textbf{Takeaway.} Many order permutations will lead to similar energy or latency performance. This is why various loop orders can be placed into large "stationarity" buckets (such as weight/ input/ output/ row)~\cite{chen2016eyeriss,maestro,timeloop} or inner/ outer product~\cite{wu2021sparseloop}.

\begin{table}[t]

\centering
\caption{MSE for workload with weight sparsity. In each columns, the blue cell shows the performance of the optimized mapping for the sparse workload; the rest of the cells shows the performnace of the same mapping tested with the workload with different sparsity. We highlight the best-performing cell of each row by green text. We can observe that the blue cells overlap with green texts, indicating that different workload with different sparsity levels do require different mapping to optimize the performance.}

\includegraphics[width=0.9\linewidth]{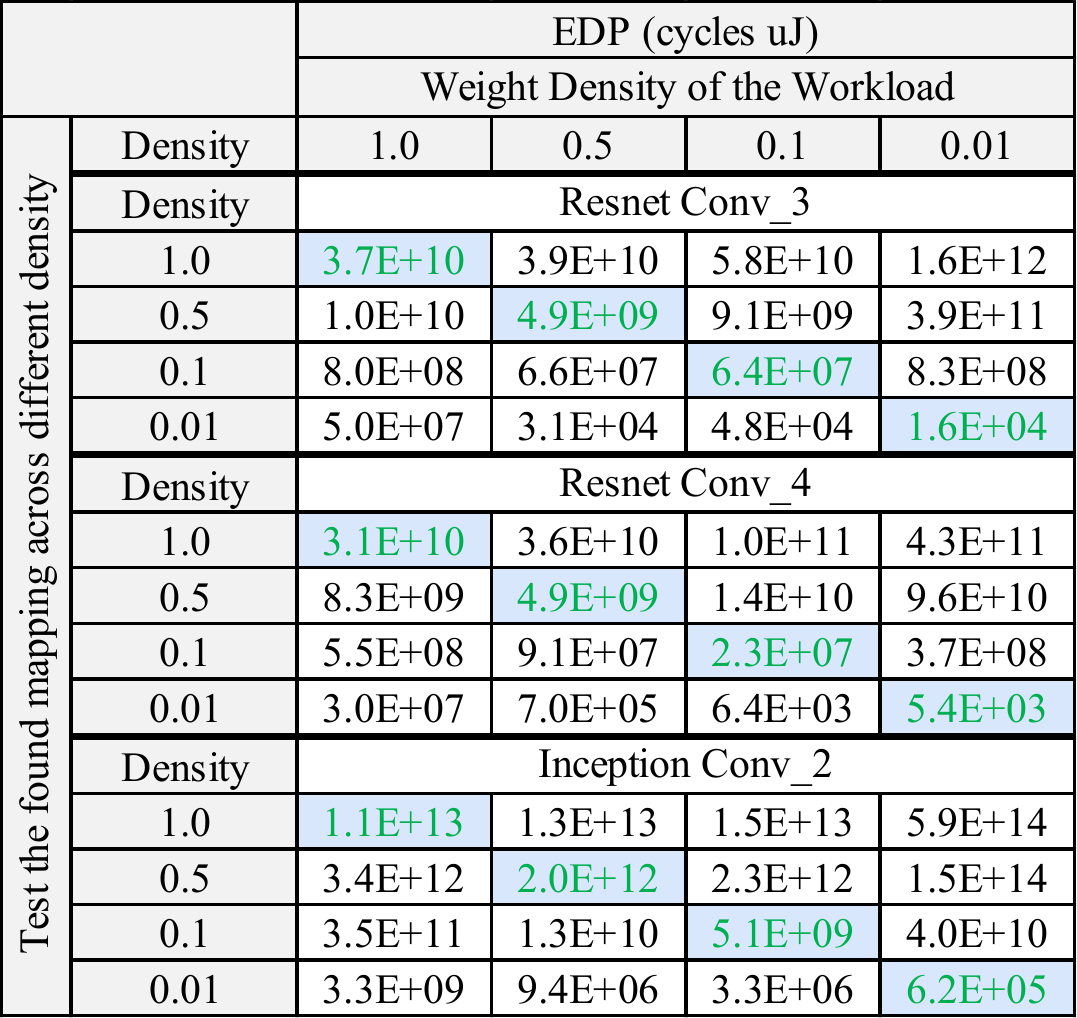}

\label{table:sparse_weight_exp}
\end{table}

\begin{table}[t]

\centering
\caption{The optimized EDP performance of inner and outer product style mapping on sparse-dense GEMM workloads in Bert-large model~\cite{transformer}. The workload density indicates the density of the sparse matrix. Bert-large KQV: the key/ query/ value projection operations. Bert-large Attn: the attention operation, Bert-large FC: the FC operations at the end of attention blocks. }

\includegraphics[width=1\linewidth]{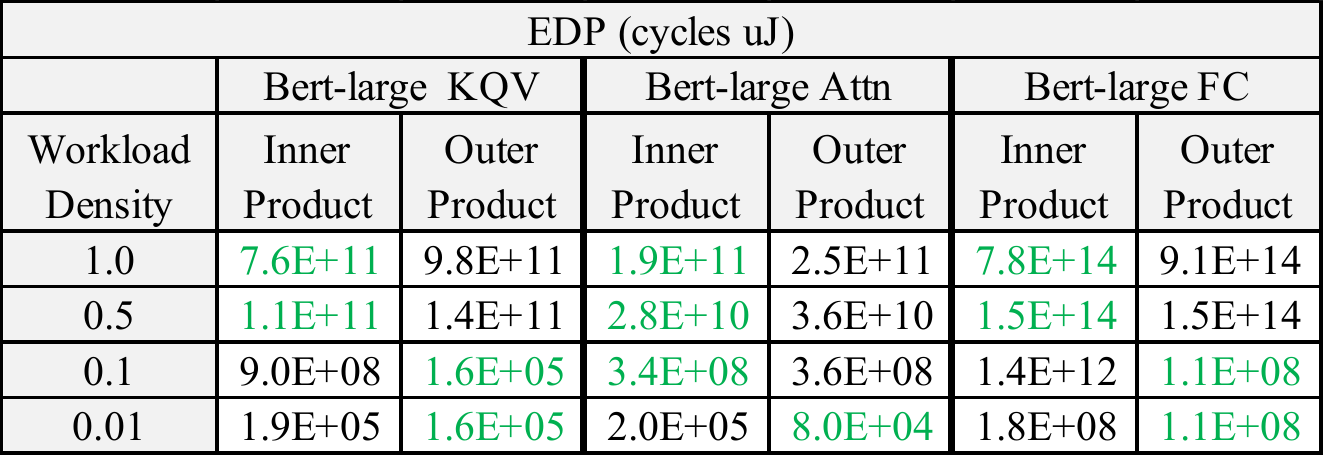}

\label{table:inner_outer_exp}
\end{table}

\subsection{Understanding Sparse Accelerator Mappings}
\label{sec:understanding_sparse}

\subsubsection{Need of MSE for Flexible Sparse Accelerator}
There is a series of research proposing ways to prune DNN models~\cite{wang2020spatten,guo2019reweighted,wang2019structured,sajjad2020poor, liu2018rethinking, li2016pruning, zhu2017prune}. However, the pruned models often cannot achieve as much performance gain in hardware as proven by the algorithmic analysis because of the increase complexity to find efficient mapping. There are several sparse accelerators~\cite{qin2020sigma,lee2018stitch,zhang2020snap,yang2019sparse,parashar2017scnn,zhang2016cambricon,kanellopoulos2019smash,kao2021e3} for efficiently running sparse workloads, skipping zeros in the weights and/or activations. However, they often employ a fixed mapping (or a limited set of mappings). 
Given the nascent domain, MSE for flexible sparse accelerators is relatively unexplored, with one study looking into it~\cite{wu2021sparseloop} in contrast to several MSE studies for flexible dense accelerators~\cite{hasco,confuciux,flexflow, flextensor,cosa,dmazerunner,ragan2013halide,vasilache2018tensor,baghdadi2019tiramisu,grosser2011polly,mindmapping,gamma,autotvm}.
This leaves MSE for sparse accelerators and workloads an area with plenty of opportunity to explore.


\subsubsection{Mapping Search for Sparse Weights}
\label{sec:weight_sparse}
For model pruning, we often focus on pruning out the weight of the models, essentially some weight becomes zero. Density 1.0 means dense weight, and density 0.5 means 50\% of the weights are zero. In \autoref{table:sparse_weight_exp}, we use workloads with different weight densities and use MSE to search for optimized mappings. The performance of found mappings are recorded in the blue cell. For example, the mapping found for Resnet CONV\_3 with 0.5 density has EDP performance of 4.9E+9 (cycles uJ). 

\textbf{Do we need different mappings for different sparsity?} We take the optimized mapping targeting a specific workload with a specific density (blue cell) and test it with the same workload with different densities. For e.g., at the top-left blue cell (\autoref{table:sparse_weight_exp}), we have an optimized mapping for the dense workload (density 1.0). Then we use the same mapping and test its performance under 0.5, 0.1, 0.01 density degrees, whose performance is recorded in the bottom cells. We perform the same experiment for the other three columns. We mark the best-performing cell across each row with green text. We can observe that the best-performing ones always located in the blue cell, meaning to optimize mapping for specific sparsity of the workload is needed to pursue the best performance.
\textbf{Takeaway.} 
A dense mapping cannot generalize across sparsity workloads. 
Different sparsity levels of the workload require different mappings to maximize the performance.

\subsubsection{Sparse Inner and Outer Product} 
\label{sec:inner_outer}
An observation that many sparse accelerators papers have made is that inner product accelerators often perform better for low sparsity workloads and outer product accelerators perform better at high amounts of sparsity~\cite{outerspace, parashar2017scnn}. We study this general observation using the MSE framework. We assume the underlying sparse accelerator is flexible to support both inner and outer product style mapping. Inner and outer products are essentially affecting the loop order. Therefore, we fix the loop order and perform MSE for the other two axes (parallelism and tile sizes). \autoref{table:inner_outer_exp} shows that the inner product style with optimized mapping consistently outperforms the outer product counterparts for workload density larger than 0.5, while the outer product style has an edge over the inner product style at densities smaller than 0.1.
\textbf{Takeaway.} From the viewpoint of MSE, we are able to validate the observation that inner product style mappings are better for denser workloads while outer product style works better at high sparsity.

\subsection{Lessons Learnt}
\vspace{-1mm}
We summarize two key takeaways from our analysis:

\begin{tcolorbox}
\squishlist
\item The feedback based mapper has the highest sampling efficiency and can directly work for any workload and accelerator configurations.
However, it has the highest wall-clock time to acquire one sample (10x more costly than random-based mappers, e.g., Random-Pruned~\cite{timeloop}). Neural architecture search is leading to new DNN models coming out frequently with highly irregular tensor shapes, increasing the demand for sample-efficient MSE.
\item MSE needs to consider sparsity. While the sparsity of the weight is often fixed for a trained DNN models, the sparsity of activations is dynamic. When facing activation sparsity, we would either under-utilize the hardware because of inefficient mapping or would need to re-launch the MSE again and again for every input-activation. 
\squishend
\end{tcolorbox}

\begin{figure*}[t]
\begin{center}
\includegraphics[width=0.87\linewidth]{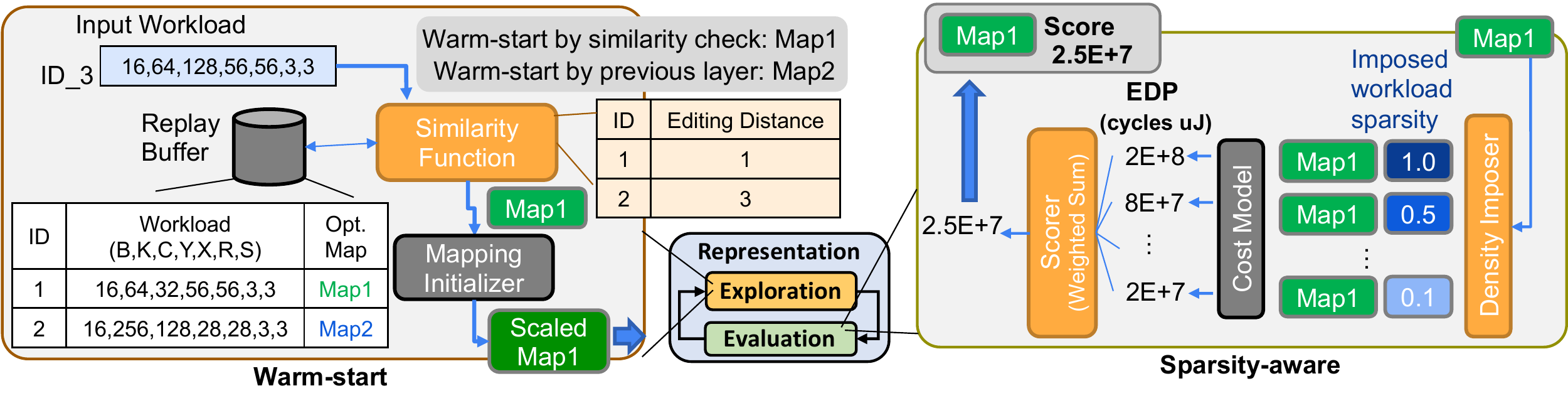}
\end{center}
\vspace{-0.60cm}
\caption{The workflow of proposed Warm-start and Sparsity-aware techniques in MSE.}
\vspace{-0.45cm}
\label{fig:techniques}
\end{figure*}

\begin{figure}[t]
\begin{center}
\includegraphics[width=0.95\linewidth]{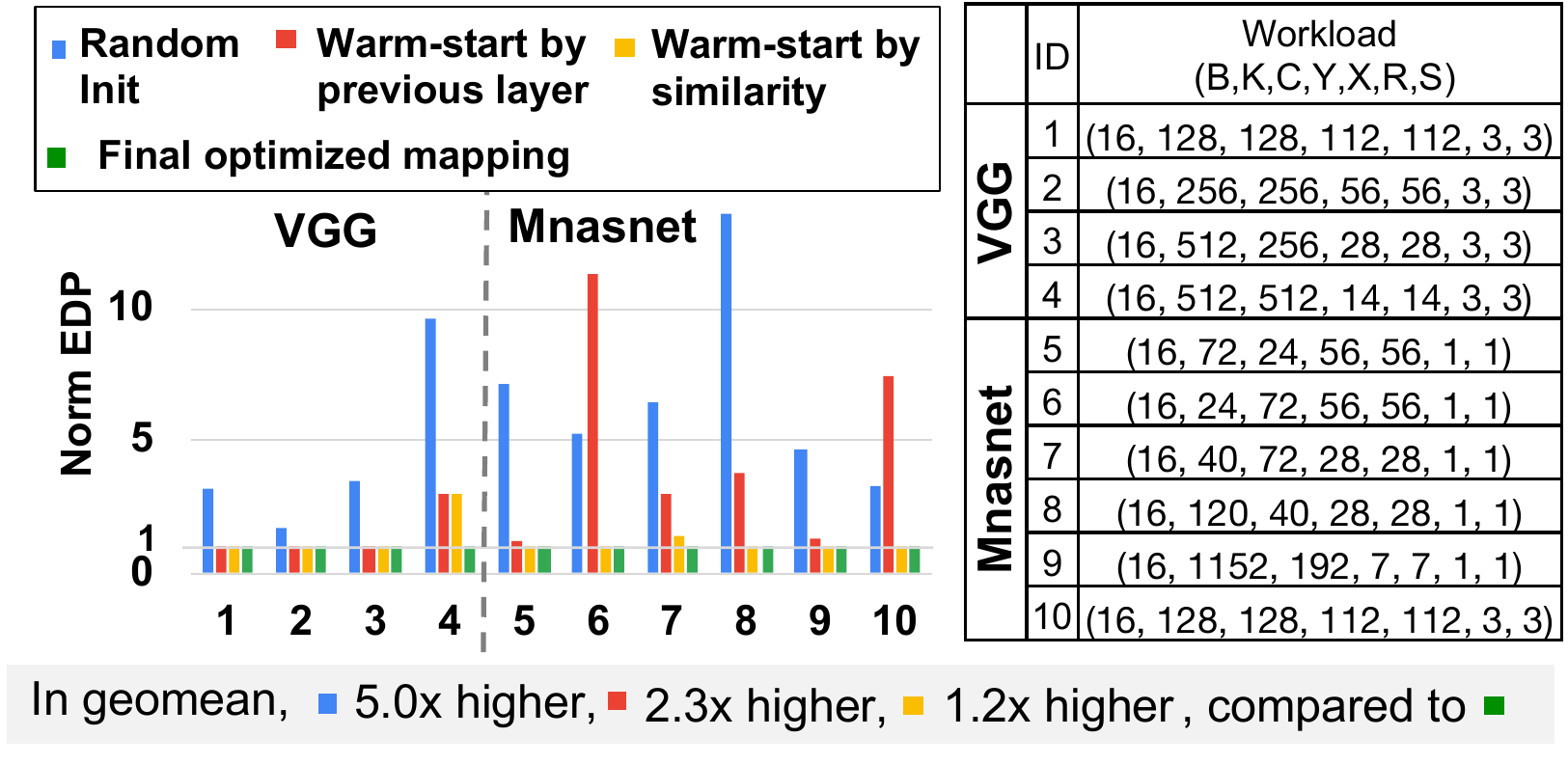}
\end{center}
\vspace{-0.50cm}
\caption{Performance comparisons of initialized solution by Random Init and two types of warm-start Init comparing to the final optimized performance (after search). The EDP values are normalized by final optimized EDP (green bars).}

\label{fig:warm_start_scaling}
\end{figure}

\begin{figure}[t]
\begin{center}
\includegraphics[width=0.9\linewidth]{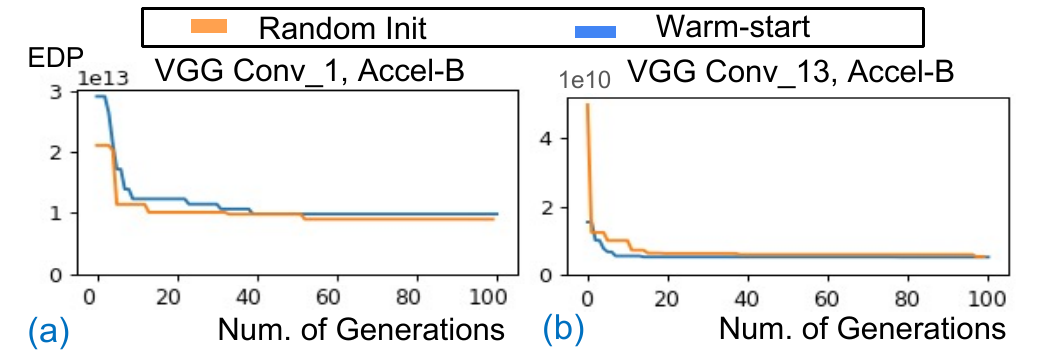}
\end{center}
\vspace{-0.40cm}
\caption{The performance convergence curve with random initialization and warm-start (by similarity) initialization at the (a) first layer and (b) a later layer of VGG16.}

\label{fig:warm_start_fig}
\end{figure}

\begin{figure}[t]
\begin{center}
\includegraphics[width=0.95\linewidth]{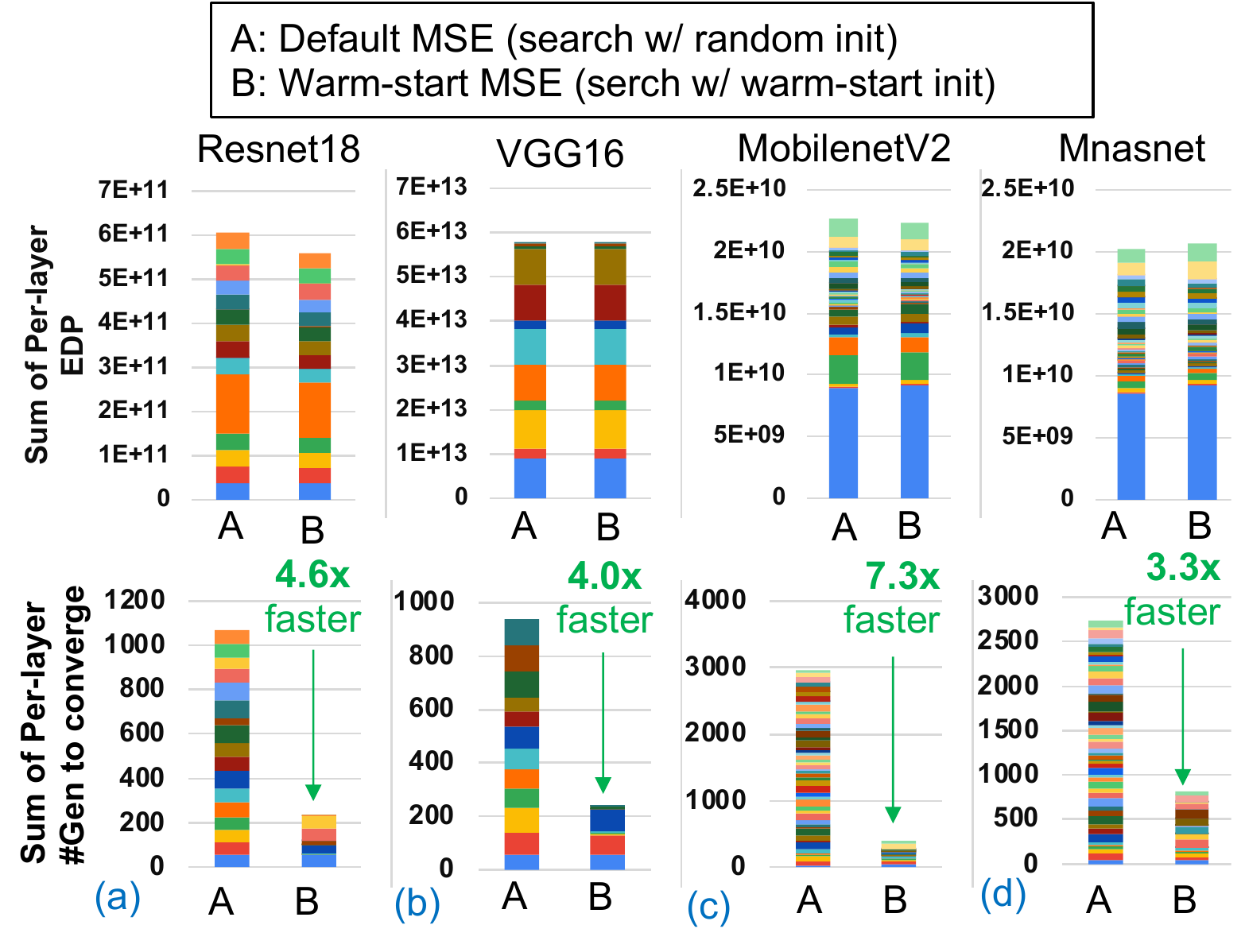}
\end{center}
\vspace{-0.40cm}
\caption{The benefit of warm-start (by similarity) when executing MSE. Warm-start MSE achieves comparable EDP performance to default MSE, but converges 3.3-7.3x faster. Different colors represent different layers of the DNN models.}

\label{fig:warm_start_bar}
\end{figure}

\section{Improving MSE}
\label{sec:algorithm}
\vspace{-1mm}
From our analysis and takeaways from \autoref{sec:analysis}, we focus on the two open-challenges identified above for next-generation mappers: search speed and sparsity.
We propose two heuristics - ``warm start" and ``sparsity-aware" to address these.

\subsection{Warm-start}
\vspace{-1mm}

\subsubsection{Motivation} We introduce \textbf{warm-start} to reduce the search time. This method is inspired by two observations. 
(1) Informed by the study in \autoref{sec:ablation} and \autoref{sec:order_study}, we know that order and parallelism are often less sensitive from workload to workload.
(2) Because of the nature of the DNN operations (CONV, FC, and others), consecutive layers often have some dimensions the same or similar to each other. Therefore potentially the mapping of the later layers can be inspired by the found mapping of the previous layer.

\subsubsection{Proposed Warm-start Search Mechanism}
\autoref{fig:techniques} shows our warm-start flow. We introduce a \emph{replay buffer} within the MSE framework which stores the optimized mapping of each workload (i.e., DNN layer) that has been run so far. We initialize the algorithm with the solution of the highest-similarity workload in the replay buffer. 

\textbf{MSE Flow.} Warm-start works via the following flow. \textit{Step-1:} When the new workload comes, we compare the workload \textit{similarity} to the workloads in the replay buffer. We use \textit{editing distance} as the similarity metric. \textit{Step-2:} Initialize the algorithm with the mapping with the highest-similarity by (i) Inherit the order and parallelism parts of the solution, and (ii) Scale the tile sizes to match the tensor dimensions of the current workload. \textit{Step-3:} Run the search algorithm.


\textbf{Walk-Through Example.} In \autoref{fig:techniques} as an example, there are two workloads that are finished with their final optimized mapping stored in the replay buffer. The next workload, workload-3, comes and will go through warm-start block before entering optimization loop. In the warm-start block, we use \textit{editing distance} to compare the similarity between the current workload and the workloads in the replay buffer. E.g., workload-3 is only differ from workload-1 in the C-dimension, leading to editing distance of 1; similarity, editing distance with workload-2 is 3 (K, Y, X). Therefore, we pick the stored optimized mapping for workload-1 (Map1), scale it to match the tensor shape of workload-3 (i.e., multiply C tile size by 2 at the outer-most tiling level (L3 mapping)), and use it as the initialized mapping for the optimization. 

\textbf{Similarity.}
Typically, for most DNNs we find that previous layer has the highest-similarity score. However, there are some exceptions: 1) the layers can come out-of-order because of other compiler decisions or 2) irregular tensor shapes of the workloads created by neural architecture search.

\subsubsection{Evaluation}

\textbf{Impact of Warm-start Initialization.}
Warm-start is an initialization technique. 
In \autoref{fig:warm_start_scaling}, we show the performance of the initialized mapping of warm-start by similarity (yellow bar), warm-start by previous layers (red bar), and the default random initialization (blue bar). We evaluate workloads from two DNN models, VGG~\cite{vgg} and Mnasnet~\cite{mnasnet}. Many DNN models are made by human experts, where the shape of each layer are often designed with high regularity such as VGG~\cite{vgg} and Resnet~\cite{resnet}. In these models, warm-start by previous layers and warm-start by similarity make no difference, since the highest-similarity layers are almost always the previous layers, as shown in workload ID 1 - 4. However, the shape of the workloads in the Mnasnet, a network found by neural architecture search, are more irregular. Therefore warm-start by similarity becomes essential, providing 2x better performance than warm-start by previous layers. However, both warm-start strategies are effective and are 2.1x and 4.3x better than random initialization.


\textbf{Impact of Warm-start Search.}
Warm-start reduces the time to converge. \autoref{fig:warm_start_fig} shows the converge curve of the first layer and a later layer to perform MSE on VGG16~\cite{vgg}. For the first layers (VGG Conv\_1), there are no previous solution in the replay buffer. Therefore, searching with random initialization or with warm-start initialization has no difference. However, for the later layers (VGG Conv\_13), searching with warm-start initialized with better points and converges faster.

We perform MSE for all layers in 4 DNN models with and without warm-start. \autoref{fig:warm_start_bar}(a) shows that searching with warm-start does not affect the quality of the found solutions, i.e., the EDP values are as low as the default algorithm. Meanwhile, warm-start can converge \textbf{3.3x-7.3x} faster (we define time-to-converge as the time to reach 99.5\% of performance improvement. In the figure we use the number of generation-to-converge, an equivalent index of time-to-converge.). We observe that Mnasnet~\cite{mnasnet} enjoys the least speedup. It is because Mnasnet is a result of neural architecture search, with irregular tensor shapes in each layer. Therefore scaling from previously-seen solutions will perform not as close to the optimized solutions as in regular networks such as Resnet~\cite{resnet}, VGG~\cite{vgg}, Mobilenet~\cite{sandler2018mobilenetv2}, which are manual designed. Nonetheless, warm-start for Mnasnet 
can still converge \textbf{3.3x} faster. 

\begin{table}[!h]

\centering
\caption{Comparisons of sparsity-aware technique and static-density heuristic when tackling the activation sparsity. The static-density heuristic searches mapping for a fixed density level (1.0, 0.5, or 0.1). At search time, the sparsity-aware technique are enabled to see the performance of a mapping on a limited sets of density levels, which are randomly picked, e.g., 1.0, 0.8, 0.5, 0.2, and 0.1 in this experiments (marked as blue cells). We highlight the best-performing one in each row with green text. Sparsity-aware will find one fixed mapping solution. We test the found mapping with a range of density (1.0 - 0.05) and record their performance. Note that many of the density levels (in 1.0 - 0.05) are never seen by MSE at search time. The result indicates that sparsity-aware technique can find mapping with comparable performance to the static-density ones across a range of sparsity.}

\includegraphics[width=0.95\linewidth]{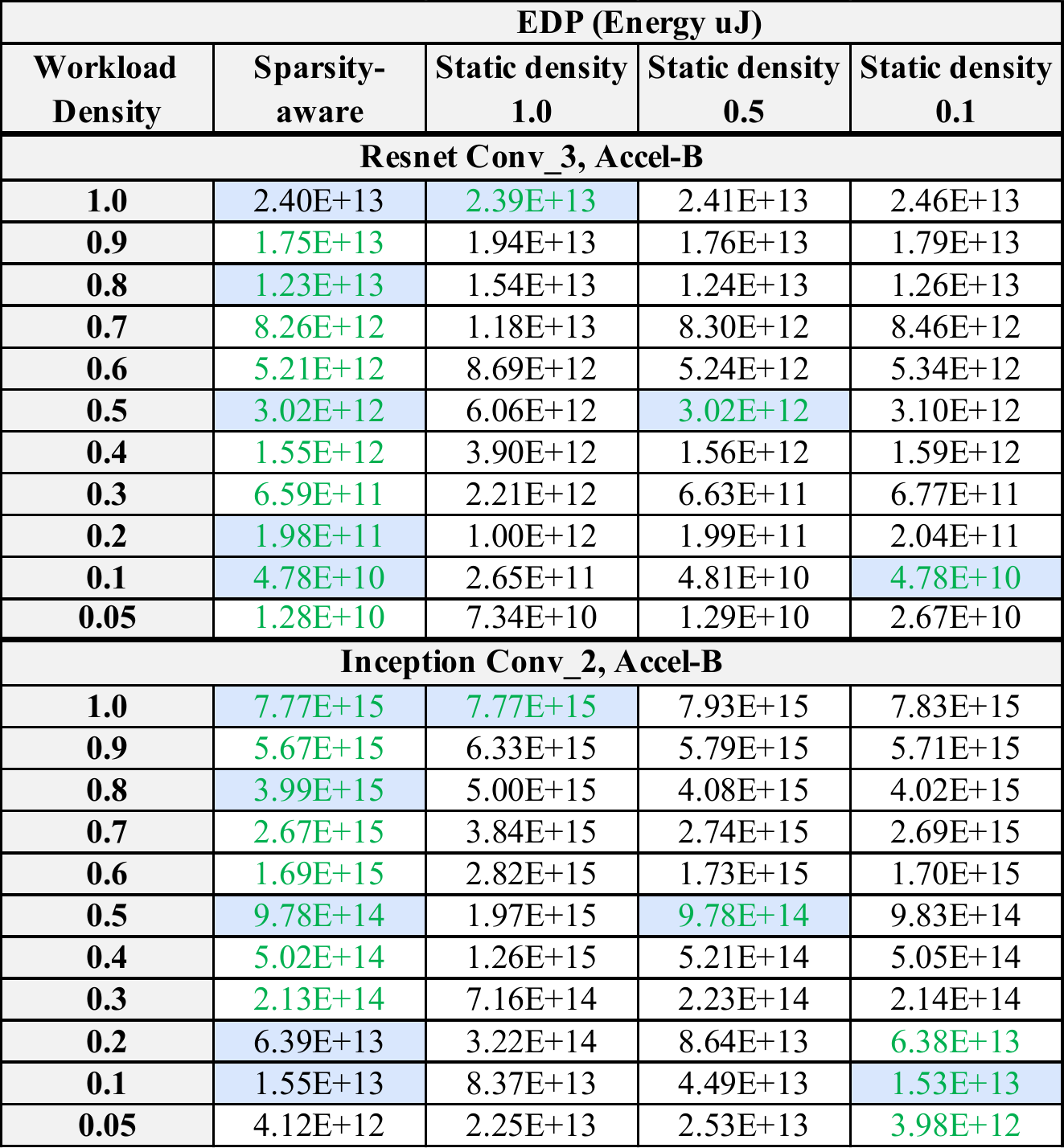}

\label{table:sparse_exp}
\end{table}

\subsection{Sparsity-aware MSE}
\vspace{-1mm}

\subsubsection{Motivation}
\label{sec:act_sparse} 
In \autoref{sec:weight_sparse} we identified the need different mappings for different sparsity of workloads. While tackling weight sparsity is straightforward because weight sparsity is often fixed at model deploy time, tackling activation sparsity is challenging. Since the activation sparsity is not known a priori before runtime, and it differs per each input data, rather than asking MSE to search for the optimal mappings for all layers and all runtime dynamic sparsity levels, we ask MSE to search for “a sparsity-aware mapping” that is efficient across a range of sparsity levels. The only information the MSE relies on is what is the typical “range” of sparsity level for a given workload, e.g., 1.0 - 0.1 for a typical DNN workload.

It is not practical to search for an optimal mapping for each new input-activation. We want to seek out \emph{if we can discover a mapping that can generalize across a range of sparsity levels to tackle the dynamic sparsity in activations?}


\subsubsection{Proposed Sparsity-aware Search Mechanism}
We propose sparsity-aware mapping search, which works as follows. When executing MSE, we don't look at the actual density level of each activation (since it is dynamic). Instead, we assume and impose sparsity in the workload when executing MSE. We impose the activation to have a density from 1.0 to 0.1, which is the typical range of activation density in DNN~\cite{qin2020sigma,lee2018stitch,zhang2020snap,yang2019sparse,parashar2017scnn,zhang2016cambricon}. Next, when executing MSE, we score the mapping by the performance of this mapping on workload across the sweep of density levels (\autoref{fig:techniques}). 

\textbf{Scoring a Mapping.} We score a mapping by the weighted sum of the performance. We use a heuristic that ``the hardware performance (e.g., latency, energy) is with positive correlation to the density of the workload'' to decide the weighting. We pick the weighting by the factor of $density$\footnote{We pick the weighting linear to $density$, since we experiment only with activation sparsity (not weight) in our evaluation.}
For example, assuming we have two density levels, 0.5 and 1.0, with hardware performance $Perf_{0.5}$ and $Perf_{1.0}$, then the (weighted sum) score is:
$\frac{Perf_{0.5}}{0.5}$ + $\frac{Perf_{1.0}}{1.0}$.


\subsubsection{Evaluation}
We compare the ``sparsity-aware'' (\autoref{sec:act_sparse}) with ``static-density'' in \autoref{table:sparse_exp}. Both ``sparsity-aware''and ``static-density'' are agnostic to the actual workload density. 
``Static-density 1.0'' always assumes the workload is dense when searching. ``Static-density 0.5'' searches the mapping assuming the workload has 0.5 density, and ``Static-density 0.1'' assumes 0.1 density. ``Sparsity-aware'' searches the mapping assuming the workload density range from 1.0 - 0.1. Specifically, we use 5 density levels: 1.0, 0.8, 0.5, 0.2, and 0.1 (blue cells in the first column), which are picked by heuristics. That is, when evaluating the mapping in the optimization loop, we scored the mapping by the performance of this mapping under workload density levels of 1.0, 0.8, 0.5, 0.2, and 0.1, and used the weighted sum of the performance as the final scores for the mapping. The scores are used to select which mappings proceed to the next iteration of the optimization loop.

We test the found mappings of the four strategies (columns) in \autoref{table:sparse_exp} by workload with density from 1.0 to 0.05. The performance of each is recorded in the corresponding rows. We make two observations: 1) The ``sparsity-aware'' can reach comparable performance to the ``static-density'' ones at the density levels, for which the ``static-densities'' are specifically optimized. For example, ``static-density 1.0'' found a mapping with EDP 2.39E+13 (cycles uJ) at density level 1.0. The mapping found by ``sparsity-aware'' can perform at a comparable EDP of 2.40E+13 (cycles uJ). 2) Aware of a range of sparsity (1.0 - 0.1), ``sparsity-aware'' can successfully find a mapping that can generalize across a range of sparsity. A fixed mapping found by ``sparsity-aware'' can achieve (in geomean) \textbf{99.7\%} of performance to the performance of each of the mappings specifically searched for different density levels.

\section{Related works}
\label{sec:relatedwork}
\vspace{-1mm}


\textbf{Map Space Exploration.}
\label{sec:mse_relatedwork}
Many mappers (search algorithms) with different algorithmic techniques are proposed to tackle the MSE problem. Timeloop-mapper~\cite{timeloop}, Simba~\cite{simba}, dmazeRunner~\cite{dmazerunner}, Interstellar~\cite{yang2020interstellar}, and others~\cite{tillet2019triton, lu2017flexflow, gao2019tangram,tetris, deeptools, suda2016throughput, shen2017maximizing, cong_fpga, scaledeep, hypar,systolic_mapping, song2018towards, stoutchinin2019optimally} use random sampling on a raw or pruned search space. Gamma~\cite{gamma}, Autotvm~\cite{autotvm}, and others~\cite{vasilache2018tensor, suda2016throughput,magma} use genetic algorithms. Tiramisu~\cite{baghdadi2019tiramisu} and Tensor Comprehensions~\cite{vasilache2018tensor} use constrained optimization. HASCO~\cite{hasco} and Reagen et. al~\cite{reagen2017case} uses Bayesian optimization, RELEASE~\cite{ahn2019reinforcement}, ConfuciuX~\cite{confuciux}, and FlexTensor~\cite{flextensor} uses reinforcement learning. 
Mind Mappings~\cite{mindmapping} 
uses a neural network-based surrogate model to replace the cost model and directly uses backpropagation to learn a solution that maximizes the objective. There are also other techniques such as  mixed-integer programming in CoSA~\cite{cosa}, MCMC search in FlexFlow~\cite{flexflow}, and others~\cite{ragan2013halide,vasilache2018tensor,baghdadi2019tiramisu,grosser2011polly}. While there have been plenty of mappers proposed, a deeper analysis of how the MSE works and how different mapping axes contribute to the performance is often lacking, which this work performs.

\vspace{-1mm}
\section{Conclusion}
\vspace{-1mm}
MSE for NPUs is a computationally expensive problem with active ongoing research. There is, however, no work, to the best of our knowledge, that has focused on understanding how different state-of-the-art mappers navigate the map-space across different axes. This work performs a deep-dive analysis on MSE using heuristic and learning-based mappers and identifies their strengths and weaknesses. We also propose two new techniques - warm-start and sparsity-aware - to enable scalability to emerging large, irregular and sparse DNNs. We hope that by our analysis, we can make MSE more approachable and understandable to a broader community, and propel the invention of advanced mapping search techniques.


\section*{Acknowledgments}
We thank Yannan Wu for the advice and support on Sparseloop setup. This work was supported in-part by NSF Award \#1909900.

\bibliographystyle{IEEEtranS}
\bibliography{main}

\end{document}